\documentclass{article}
\usepackage[english]{babel}
   
\usepackage{amsfonts}
\usepackage{float}
\usepackage{color}
\usepackage[hyphens]{url} 
\usepackage{ragged2e}
\usepackage{hyperref}
\hypersetup{
    colorlinks=true,
    linkcolor=blue,
    filecolor=magenta,      
    urlcolor=cyan,
    pdftitle={Overleaf Example},
    pdfpagemode=FullScreen,
    }
\usepackage[export]{adjustbox}
\usepackage{xcolor}

\usepackage[page]{appendix}
\usepackage{amsmath}
\usepackage{graphicx}

 \usepackage{stfloats}
 
\usepackage{microtype}
\usepackage{graphicx}
\usepackage{subfigure}
\usepackage{booktabs} 

\usepackage[accepted]{icml2024}

\usepackage{amsmath}
\usepackage{amssymb}
\usepackage{mathtools}
\usepackage{amsthm}

\usepackage[capitalize,noabbrev]{cleveref}

\theoremstyle{plain}

\theoremstyle{definition}

\theoremstyle{remark}

\usepackage[textsize=tiny]{todonotes}

\icmltitlerunning{Investigating the Indirect Object Identification circuit in Mamba}

\begin{document}

\twocolumn[
\icmltitle{Investigating the Indirect Object Identification circuit in Mamba}



\icmlsetsymbol{equal}{*}

\begin{icmlauthorlist}
\icmlauthor{Danielle Ensign}{ind}
\icmlauthor{Adrià Garriga-Alonso}{far}
\end{icmlauthorlist}

\icmlaffiliation{ind}{Independent}
\icmlaffiliation{far}{FAR AI}

\icmlkeywords{Machine Learning, Mechanistic Interpretability, Mamba, State Space Models, Large Language Models, ICML}

\vskip 0.3in
]
\printAffiliationsAndNotice{}

\begin{abstract}
How well will current interpretability techniques generalize to future models? A relevant case study is Mamba, a recent recurrent architecture with scaling comparable to Transformers. We adapt pre-Mamba techniques to Mamba and partially reverse-engineer the circuit responsible for the Indirect Object Identification (IOI) task. Our techniques provide evidence that 1) Layer 39 is a key bottleneck, 2) Convolutions in layer 39 shift names one position forward, and 3) The name entities are stored linearly in Layer 39's SSM. Finally, we adapt an automatic circuit discovery tool, positional Edge Attribution Patching, to identify a Mamba IOI circuit. Our contributions provide initial evidence that circuit-based mechanistic interpretability tools work well for the Mamba architecture.

\end{abstract}

\section{Introduction}

If we care about using interpretability on new models, we should know:
Will interpretability techniques generalize to new architectures? 

To investigate this question, we apply existing mechanistic interpretability techniques to a new model developed after most interpretability techniques: Mamba. Mamba is a State Space Model (SSM), a type of recurrent neural network \citep{gu2023mamba}. Mamba is the result of years of work on language modeling with state space models \citep{gu2020hippo,gu2022efficiently,fu2023hungry}, and is one of many new RNN-like architectures \citep{beck2024xlstmextendedlongshortterm,peng2023rwkvreinventingrnnstransformer,gu2023mamba,lieber2024jamba}. These RNNs have scaling competitive with Transformers, unlike LSTMs \citep{scalinglaws}. Because it is a recurrent network, it only needs to store hidden states from the previous token, resulting in faster inference. The recurrence is also linear (and thus associative) over token position, which permits further optimizations. See \hyperref[mamba-description]{Appendix} for architecture details.

While we are the first to focus on finding circuits in Mamba, previous work has shown other interpretability techniques apply. For example, \citet{sharma2024locating}
locate and edit factual information with ROME (Rank One Model Editing) \citet{meng2023locatingeditingfactualassociations}. \citet{ali2024hidden} extract hidden attention matrices, and \citet{othello_mamba, grazzi2024mamba} use linear probes to identify capabilities. Additionally, \citet{paulo2024does} showed that Contrastive Activation Addition \citep{rimsky2024steering}, Tuned Lens \citep{belrose2023eliciting} and probes to elicit latent knowledge \citep{mallen2024eliciting} transfer to the Mamba architecture.

This work focuses on applying techniques from circuit-based mechanistic interpretability to Mamba to see how well these techniques transfer to new architectures. In particular, we study \href{https://huggingface.co/state-spaces/mamba-370m}{$\text{state-spaces/mamba-370m}$}, a 370-million-parameter Mamba model pretrained \citep{gu2023mamba} on The Pile \citep{pile}. We chose this model as it is the smallest Mamba model with good performance ($\sim 96\%$ accuracy on our templates) on the Indirect Object Identification (IOI) task \citep{wang2022interpretability}.

In particular, for the IOI task, we:
\begin{enumerate}
    \item Show multiple lines of evidence suggesting layer 39 is a bottleneck:
    \begin{enumerate}
        \item Zero and Resample ablation (Section~\ref{sec:ablations}) experiments point to layer 39 and layer 0.
        \item We compute greedy minimal subsets of layers allowed to transfer information across tokens (``token cross-talk''). These always include layer 39 (but layer 0 only 18\% of the time).
    \end{enumerate}
    \item Provide evidence that the convolution on layer 39 shifts name data to the next token position.
    \item Modify the representations used by layer 39 using averages of activations, resulting in overwriting one output name with another \citep{rimsky2024steering}. These results suggest that in the SSM of layer 39, entity names are linearly represented, with different representations for the first and second time (or sentence) the names appear in IOI.
    \item Provide multiple lines of evidence that layer 39 writes outputs into only the final token position:
    \begin{enumerate}
        \item Resample ablation on the hidden state and the values added to the residual stream.
        \item A slight modification of the results of EAP gives us a subgraph that is capable of doing IOI while only leaving layer 39's final token's outputs unpatched.
    \end{enumerate}
\end{enumerate}

\begin{figure}
    \centering
    \includegraphics[width=0.9\linewidth]{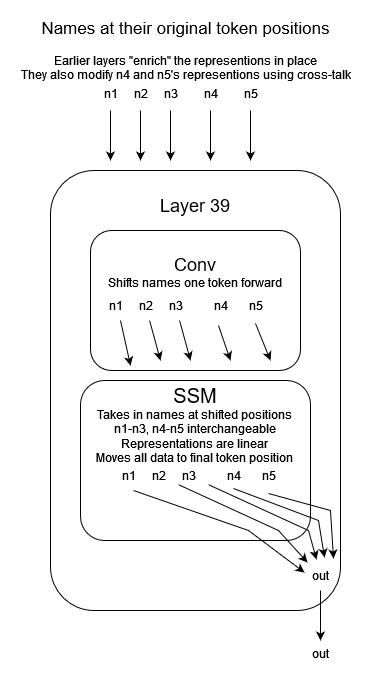}
    \caption{Our hypothesis for the role of Layer 39. The representations of n1--n3 and n4--n5 are interchangeable over positions.}
    \label{hypothesisfigure}
\end{figure}

In addition, we show that ACDC and Edge Attribution Patching \citep{syed2023attribution} both result in sparse graphs when applied to IOI on Mamba, and provide the resulting computational graphs.

\section{The Test Subject: Mamba}\label{mamba-description}

Here we provide a brief overview of the Mamba architecture. We refer the reader to \citet{ophiology} for a more detailed introduction. We use $\stackrel{[A, B]}{v}$ to denote that variable $v$ has shape $[A, B]$.

\subsection{State Space Model (SSM)}

Mamba's SSM block can be written as mapping a 1D space to a N-dimensional state space, then back to a 1D space:

\begin{align}
\stackrel{[N]}{h_{t}} &= \stackrel{[N,N]}{\overline{A}}\stackrel{[N]}{h_{t-1}} + \stackrel{[N,1]}{\overline{B}}\stackrel{[1]}{x_{t}} \\
\stackrel{[1]}{y_t} &= \stackrel{[1,N]}{C}\stackrel{[N]}{h_t} + \stackrel{[1,1]} D\stackrel{[1]} x_t
\end{align}

In Mamba $A$ is diagonal, so we will just write it as $\stackrel{[N]}{\overline{A}}$ and do an element-wise product ``$\odot$''.

Each layer does $E$ of these in parallel. $A$ has a separate value for each $e$, and is encoded as an $[E,N]$ matrix. We can denote $\overline{A_e}$ as the $N$-sized entry for stream $e$, giving us,

\begin{align}
\stackrel{[N]}{h_{t,e}} &= \stackrel{[N]}{\overline{A_e}}\odot\stackrel{[N]}{h_{t-1,e}} + \stackrel{[N,1]}{\overline{B}}\stackrel{[1]}{x_{t,e}} \\
\stackrel{[1]}{y_{t,e}} &= \stackrel{[1,N]}{C}\stackrel{[N]}{h_{t,e}} + \stackrel{[1,1]} D\stackrel{[1]} x_{t,e}
\end{align}

Finally, $\overline{A}_e$, $\overline{B}$, and $C$ depend on the SSM input, and so gain a subscript $t$. $\overline{B}$ also gains a subscript $e$ through the variable time-step, $\stackrel{[1]}{\Delta_{t, e}}$. The final SSM expressions are:

\begin{align}
\stackrel{[N]}{h_{t,e}} &= \stackrel{[N]}{\overline{A_{t,e}}}\odot\stackrel{[N]}{h_{t-1,e}} + \stackrel{[N,1]}{\overline{B_{t,e}}}\stackrel{[1]}{x_{t,e}}\\
\stackrel{[1]}{y_{t,e}} &= \stackrel{[1,N]}{C_t}\stackrel{[N]}{h_{t,e}} + \stackrel{[1,1]} D\stackrel{[1]} x_{t,e},
\end{align}
where
\begin{align}
    \stackrel{[N]}{\overline{A_{t,e}}} &= \exp(-\stackrel{[1]}{\Delta_{t,e}}\exp(A_{\text{log}})_e), \\
    \stackrel{[N]}{\overline{B_{t,e}}} &= \stackrel{[1]}{\Delta_{t,e}}\stackrel{[N]}{B_{t}}, \quad \quad\text{with} \stackrel{[N]},{B_{t}} = \stackrel{[N,E]}{W_B}\stackrel{[E]}{x_t}, \\
\stackrel{[N]}{C_t} &= \stackrel{[N,E]}{W_C}\stackrel{[E]}{x_t}, \\
\stackrel{[1]}{\Delta_{t,e}} &= \text{softplus}(\stackrel{[E]}{x_{t}} \cdot \stackrel{[E]}{W^{\Delta}_e} + \stackrel{[1]}{B^{\Delta}_e}),
\end{align}

with $\stackrel{[E,E]}{W^{\Delta}}, \stackrel{[E]}{B^{\Delta}}, \stackrel{[N,E]}{W_B}, \stackrel{[N,E]}{W_C}, \stackrel{[N,E]}{A_{\text{log}}}$ being learned parameters, and $\text{softplus}(x) = \log(1+e^{x})$. This parameterization guarantees that $\overline{A}<1$, and thus the hidden state does not explode.

\subsection{Architecture}

Mamba has multiple layers which each add to a residual stream. Each layer does:

\begin{itemize}
\item Project input $\overset{[B,L,D]}{\text{resid}}$ to $\overset{[B,L,E]}{x}$
\item Project input $\overset{[B,L,D]}{\text{resid}}$ to $\overset{[B,L,E]}{\text{skip}}$
\item Conv over the time dimension, with a different filter for each $e\in [E]$ ($x = \text{conv}(x)$)
\item Apply non-linearity (silu) ($x = \text{silu}(x)$)
\item $y = SSM(x)$
\item Gating: $y = y * \text{silu}(\text{skip})$
\item Project $\overset{[B,L,E]}{y}$ to $\overset{[B,L,D]}{\text{output}}$
\end{itemize}

\begin{figure}
    \centering
    \includegraphics[width=1\linewidth]{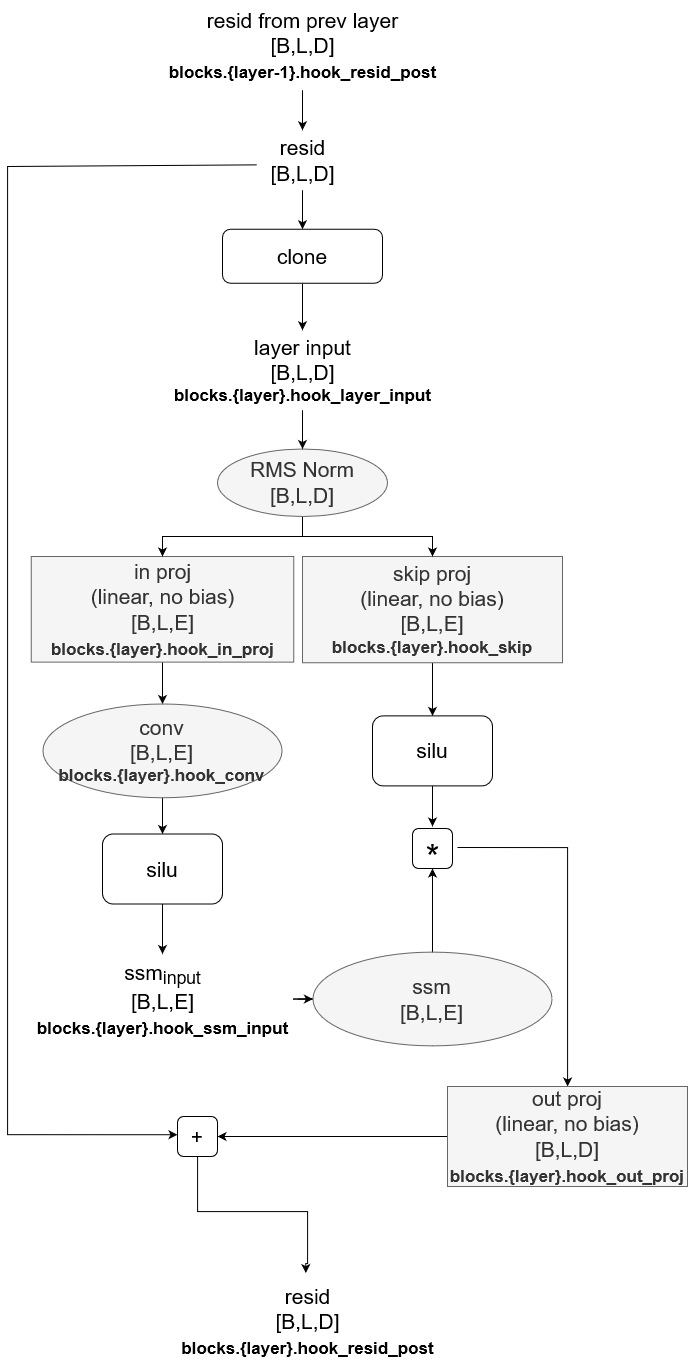}
    \caption{A single layer in the Mamba architecture, with hook points listed in all the locations we intervene. Note that the SSM contains further hook points, described in Section~\ref{overwritename}, ``\hyperref[overwritename]{Controlling model output}''.
    The ``SSM'' and ``conv'' components are affected by previous time steps.}
    \label{fig:mamba-hooks}
\end{figure}

Where $B$ is batch size, $L$ is context length, $D$ is embedding dimension, and $\text{silu}(x)=x * \text{sigmoid}(x).$ See Figure~\ref{fig:mamba-hooks}.

\section{Circuit-based Mechanistic Interpretability}


To understand how large language models (LLMs) implement their emergent capabilities \citep{emergent},
we focus on finding human-interpretable algorithms \citep{AnthropicMechanisticEssay}. This involves representing models as computational graphs and identifying circuits that are subsets of that computational graph. Ideally, each subgraph would also be annotated to describe the role of each component \citep{causal_abstraction_geiger}.

Finding circuits that capture the behavior on all inputs is intractable for large language models. Therefore, we study behavior on specific tasks.

\subsection{Problem Description}

Following \citep{causal_abstraction_geiger,conmy2023automated}:
we have a behavior (task) that we would like to study, a metric for evaluating performance, and a coarse-grained computational graph of the neural network on which we express explanations. We would like to find the minimal subgraph that attains a high enough metric score (where target metric score is a hyperparameter), with an explanation of what variations in the data each graph component captures.

\subsubsection{IOI Task}

We are studying the IOI task, initially examined by \citep{wang2022interpretability}. Consider an example data point:

\begin{verbatim}
Friends Isaac, Lucas and Lauren went to
the office. Lauren and Isaac gave a
necklace to 
\end{verbatim}

The model is asked to predict the next token, and the correct answer is ``\texttt{ Lucas}''. ``\texttt{ Lucas}'' is the Indirect Object we are trying to identify. See the \hyperref[appendix-ioi]{Appendix} for detailed data-generation templates and corruption information.

\subsubsection{Metric}\label{metric}

There are many choices of metrics: KL-Divergence, Logit Diff, Accuracy, Probability of the correct answer, etc. 
The best metric to use in general is an open question and may be task specific. For IOI, \citet{zhang2024best} suggest the Normalized Logit Diff metric, as that helps propagate information missed by accuracy. See the \hyperref[appendix-metric]{Appendix} for further details.

\subsubsection{Computational Graph}

We use the \hyperlink{https://github.com/phylliida/mambalens}{MambaLens} library (see also \citet{nanda2022transformerlens}) to intervene at different locations per experiment.

\begin{flushleft}
\begin{itemize}
    \item Section ~\ref{resampleablation} ``\hyperref[resampleablation]{Resample Ablation}'' uses \texttt{blocks.\{layer\}.hook\_layer\_input}.
    \item Section ~\ref{layerremoval} ``\hyperref[layerremoval]{Layer Removal}'' uses \texttt{blocks.\{layer\}.hook\_out\_proj}.
    \item Section ~\ref{crosstalk} ``\hyperref[crosstalk]{Removing Cross Talk}'' uses \texttt{blocks.\{layer\}.hook\_in\_proj}.
    \item Section ~\ref{layer39convshift} ``\hyperref[layer39convshift]{Layer 39 Uses Conv to Shift Names One Position Forward}'' uses \texttt{blocks.\{layer\}.hook\_in\_proj} and \texttt{blocks.\{layer\}.hook\_conv}.
    \item Section ~\ref{overwritename} ``\hyperref[overwritename]{Controlling model output by modifying representations on layer 39}'' uses \texttt{blocks.\{layer\}.hook\_ssm\_input}. For the cosine sim plots, it uses further hooks inside the ssm, described in \hyperref[appendix-ssm-hooks]{Appendix}.
    \item Section ~\ref{layer39moveintolasttokenpos} ``\hyperref[layer39moveintolasttokenpos]{Layer 39 moves information into only the last token position}'' uses \texttt{blocks.\{layer\}.hook\_h.\{token\_pos\}}, \texttt{blocks.\{layer\}.hook\_out\_proj}, and the hooks used in Section ~\ref{appendix-eap}
    \item Both Section ~\ref{appendix-eap} ``\hyperref[appendix-eap]{EAP}'' runs use \texttt{blocks.\{layer\}.hook\_layer\_input} and \texttt{blocks.\{layer\}.hook\_layer\_output}. They also use \texttt{hook\_embed} (described the next section) as the input node, and \texttt{blocks.47.hook\_resid\_post} as the output node.
    \item The ``\hyperref[ACDC]{ACDC}'' (Section ~\ref{ACDC}) run following the (non positional) EAP run uses all the hooks from the EAP runs. It also uses \texttt{blocks.\{layer\}.hook\_skip}, \texttt{blocks.\{layer\}.hook\_conv}, and \texttt{blocks.\{layer\}.hook\_ssm\_input}.
\end{itemize}
\end{flushleft}
Note that \texttt{blocks.\{layer\}.hook\_layer\_input} is the residual stream before normalization. If we patched directly on these, it would modify downstream values as well. Thus, to patch only a single layer's input, we clone this value first.

\subsubsection{Ablations\label{sec:ablations}}

To identify which nodes and edges are important, we take inspiration from causal inference \citep{pearl_causality}: ablate nodes of our computational graph and observe changes in the output.

Replacing activations with zero \citep{olsson2022context,cammarata2021curve} or the mean over many data points \citep{wang2022interpretability} was initially used. However, these can result in activations that are out of distribution \citep{causal_scrubbing}. \emph{Resample ablation} \citep{causal_abstraction_geiger}, also known as \emph{interchange interventions} and \emph{causal tracing}, is a commonly used alternative \citep{greaterthan,docstring,wang2022interpretability,conmy2023automated}. Resample ablation begins by running a forward pass with a \emph{corrupted} prompt, then substitutes those corrupted activations into a forward pass run on the uncorrupted prompt.

\begin{figure}
    \centering
    \includegraphics[width=0.5\linewidth]{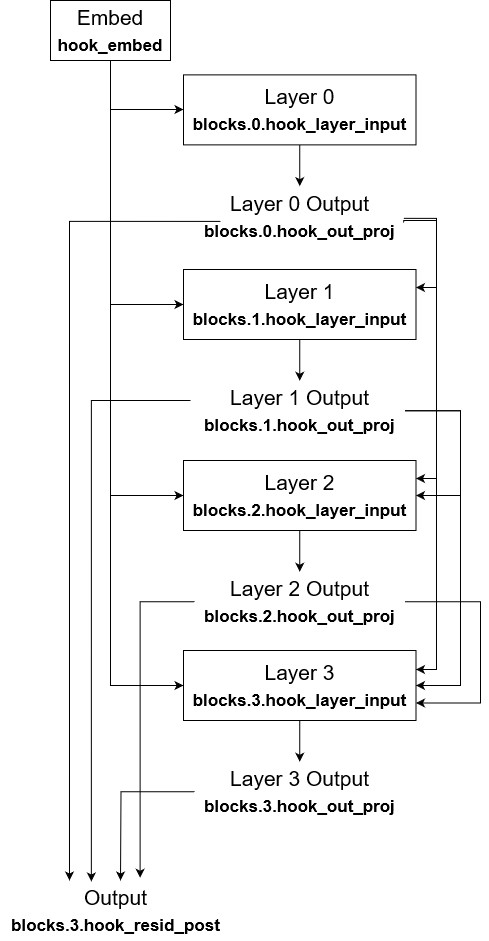}
    \caption{Fully connected causal graph, using the additivity of the residual stream. This is an example network with 4 layers, so the output node is \texttt{blocks.3.hook\_resid\_post}. The full network we study has 48 layers, so the output node is \texttt{blocks.47.hook\_resid\_post}}
    \label{everylayeredgefigure}
\end{figure}

In addition to resample ablation, in Mamba (and Transformers), the residual stream is a sum of outputs from every layer. This allows us to create an edge between every layer \citep{elhage2021mathematical}, see Figure \ref{everylayeredgefigure}.

To patch an edge this causal graph \citep{elhage2021mathematical} going from layer $i$ to layer $j$, we can do:

\begin{align}
\begin{aligned}
\overset{[B,L,D]}{\text{patched input}_j}=& \overset{[B,L,D]}{\text{input}_j} - \overset{[B,L,D]}{\text{output}_i} + \overset{[B,L,D]}{\text{corrupted output}_i}
\end{aligned}
\end{align}

We give the dimensions of our tensors in square brackets [ ] above the term. Here, $B$ is batch size, $L$ is context length, and $D$ is embedding dimension. The $\text{output}_i$ is \texttt{blocks.i.hook\_out\_proj}, computed during the same forward pass as the  $\text{input}_j$ and $\text{patched input}_j$ (which both use \texttt{blocks.j.hook\_layer\_input}). $\text{corrupted output}_i$ are stored values from a separate forward pass using the corrupted prompt and \texttt{blocks.i.hook\_out\_proj}.
    
\subsection{Semi-automatic Circuit Discovery}

Initially, finding circuits had to be done by hand: patching subsets of nodes and edges (known as path patching \citep{goldowskydill2023localizing}) until a circuit emerges. Several methods have since been developed to automate this process. Subnetwork probing learns a mask over the graph using gradient descent \citep{subnetwork_probing}. Automated Circuit DisCovery (ACDC) starts from sink nodes and works backwards to reconstruct the causal graph \citep{conmy2023automated}. ACDC requires a separate forward pass for every edge, and this can be very time-consuming. Head Importance Score for Pruning \citep{sixteen_heads}, and more recently EAP (Edge Attribution Patching) \citep{syed2023attribution}, use the gradient to approximate the contribution of all edges simultaneously. In particular, EAP approximates the \emph{attribution scores} of an edge between layer $i$ and layer $j$ via:

\begin{align}
\begin{aligned}
\overset{[B,L,D]}{\text{attr}_{i\mapsto j}} = (\overset{[B,L,D]}{-\text{output}_i} + \overset{[B,L,D]}{\text{corrupted output}_i}) \nabla \overset{[B,L,D]}{\text{input}_j}
\end{aligned}\label{attr_eq}
\end{align}

Where $\nabla \text{input}_j$ is the gradient given from the backward hook made in these steps:
\begin{enumerate}
    \item For every layer, create a backward hook on \texttt{blocks.{j}.hook\_layer\_input}
    \item Run a \texttt{forward} pass that patches every edge
    \item Compute the metric on the resulting logits, and call \texttt{backward} on the metric's value.
\end{enumerate}

To get an attribution for each edge, we \texttt{sum} $\text{attr}_{i\mapsto j}$ over the $L$ and $D$ axes, then \texttt{mean} over the $B$ axis.

This approximation can be improved by using integrated gradients \citep{marks2024sparse, sundararajan2017axiomatic}: compute a separate $\overset{[B,L,D]}{\text{attr}_{i\mapsto j}}(\overset{[1]}{\alpha_k})$ for an $\overset{[1]}{\alpha_k}=k/(\text{ITERS}-1)$ where $k \in [0,\hdots,\text{ITERS}-1]$, then compute the average of all these scores (\text{ITERS} is an int hyperparameter that determines how fine grained our approximation is, usually 5-10 is large enough). The attribution is computed in using Equation \ref{attr_eq} like before, however, the forward pass for a given $\alpha_k$ only ``partially'' applies every patch as follows:

\begin{align}
\begin{aligned}
\overset{[B,L,D]}{\text{patched input}_j} = \overset{[B,L,D]}{\text{input}_j} - \overset{[1]}{\alpha_k}(\overset{[B,L,D]}{\text{output}_i} + \overset{[B,L,D]}{\text{corrupted output}_i})
\end{aligned}
\end{align}

Once we have these attribution scores, we can sort all edges by their attribution and perform a binary search to find the minimal set of edges that achieves our desired metric.

The major downside of these automated methods is that (aside from token-level attributions) they do not yet assign interpretations to nodes.

\section{Findings}
\subsection{Layer 39 is Important}

We have three lines of evidence suggesting layer 39 is important. While two of these lines of evidence also suggest layer 0 is important, we also provide evidence that token cross-talk in layer 0 is not usually needed.

\subsubsection{Resample Ablation}\label{resampleablation}

To determine which layers are important, we will resample ablate \texttt{blocks.\{layer\}.hook\_layer\_input}.
To determine which tokens matter, we do this patch separately for each (layer, token\_position) pair. This forces us to limit to the \hyperref[appendix-ioi]{three templates} that share name token positions (one could use more templates and use semantic labels instead of token positions, but that is left to future work).

Because \hyperref[appendix-ioi]{each corruption} affects different positions, averaging over them does not make sense. Thus, we show results separately for each corruption. We focus on 3-name templates. While 2-name templates are simpler, we find results from 2-name templates to be misleading as the task is too simple.

\begin{figure*}[hbtp]
    \centering
    \includegraphics[width=0.7\linewidth]{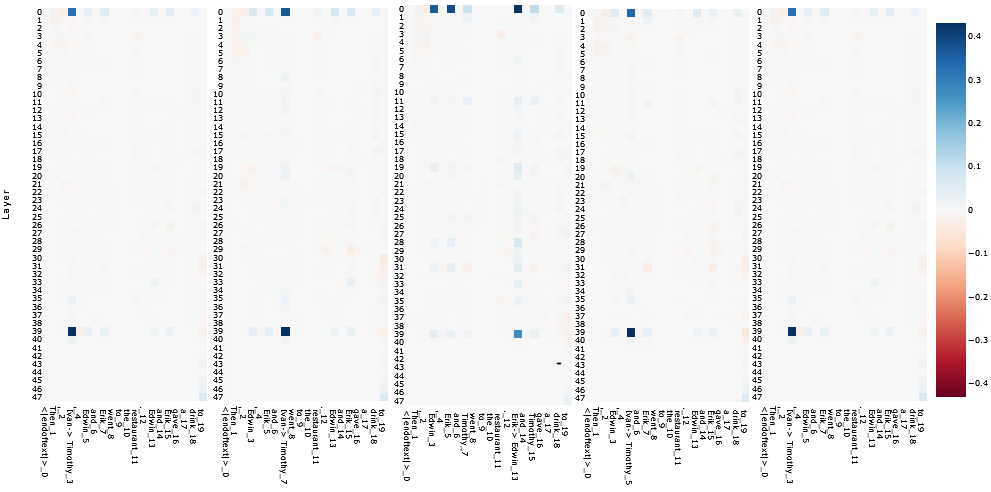}
    \caption{ Displayed is 1 - (Normalized logit diff) for each (layer, position) patch, averaged over 80 data points. 0 corresponds to acting like the uncorrupted forward pass, and 1 corresponds to acting like the corrupted forward pass. The y-axis is Layer, and the x-axis is token position. The corruptions can be observed by inspecting the token position labels. Each of the five plots correspond to different IOI patches.}
    \label{resampleablationfigure}
\end{figure*}

Figure \ref{resampleablationfigure} shows that normalized logit diff changes most when patching layer 0 and 39.

\subsubsection{Layer Removal}\label{layerremoval}

Each layer adds to the residual stream. This allows us to ``remove'' a layer by setting this added value to zero, i.e., zero-ablating layer outputs (\texttt{blocks.\{layer\}.hook\_proj\_out}). We plot probability of the correct answer, as there is no corrupted answer to compare to.

\begin{figure}
    \centering
    \includegraphics[width=1\linewidth]{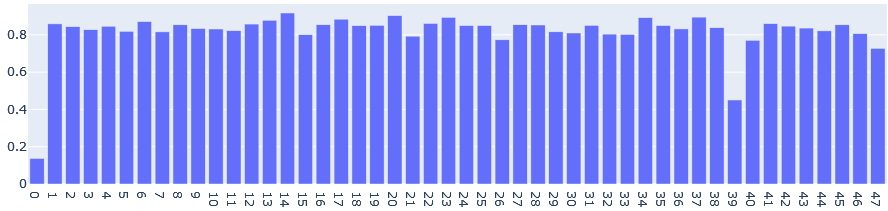}
    \caption{Relative probability of the correct token when zero-ablating each layer's outputs. Relative probability is the softmax over the 4 logits from prompt and corruption names. The clean model gets 83\%.}
    \label{layerremovalfigure}
\end{figure}

In Figure \ref{layerremovalfigure}, we again see that, layers 0 and 39 are crucial parts of the circuit that cannot be removed.

We also find that by repeatedly removing the layer that decreases accuracy the least, about half of the layers can be removed with minimal impact on accuracy. We replicated this layer removal robustness on GPT-2-Small \citep{radford2019language}. This might be seen as evidence for the residual stream having a privileged basis that is consistent between layers. However, it is also consistent with there being multiple distinct spaces (for example, embed-0, 0-39, 39-out), or layers being simultaneously compatible with multiple different spaces. See \citet{belrose2023eliciting} for more discussion on this ``privileged basis'' perspective.

\subsubsection{Removing Token Cross-Talk}\label{crosstalk}

It would be useful to know where information travels between tokens, as opposed to just modifying the representations in place. We conduct an experiment to find a small set of layers that do this ``token cross-talk''.

There are two ways in which a layer at a specific token position can affect future positions (``token cross-talk''): the convolutional (conv) layer, and the SSM block. (For clarity, in transformers, attention is where ``token cross-talk'' occurs, as that is where information can flow between different token positions)

Putting corrupted data into the conv will also put corrupted data into the SSM, as it is downstream of the conv. Thus, to remove a specific layer's ability to have token cross-talk, we can apply resample ablation to that layer's conv inputs (\texttt{blocks.\{layer\}.hook\_in\_proj}) at all positions. Because we also patched convs in previous positions, the SSM will only have information about the corrupted input. 

If we patch every layer before $L$ in the manner above, this removes any information about previous tokens at layer $L$. However, if we only patch some previous layers, the previous tokens can have influence: a previous layer could move two tokens into the same position, and then a later layer could process those token interactions in place.

Also, note that this does not completely remove ``cross talk'', it only removes cross talk that is specific to the uncorrupted prompt (i.e., cross talk that is needed for outputting the correct answer). Cross talk that occurs in both uncorrupted and corrupted prompts will still occur. This is somewhat acceptable because we only care about task-relevant cross talk.

Given these two disclaimers, we still feel this is a useful proxy for ``removing cross talk''.

Now, start with patching all layers' cross talk, then ``unpatch'' the layer that improves accuracy the most. This is repeated until accuracy is about 0.9, resulting in a ``minimal cross talk circuit'' that can perform the task. We do this separately for each (corruption, template) \hyperref[appendix-ioi]{pair}.

\begin{figure}
    \centering
    \includegraphics[width=1\linewidth]{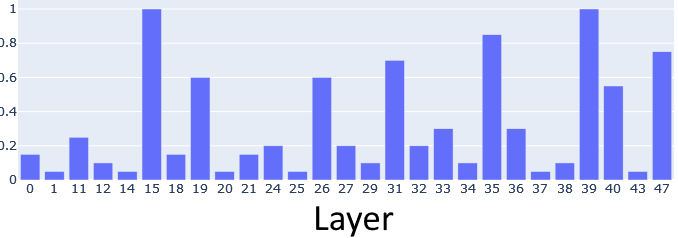}
    \caption{Out of all \hyperref[appendix-ioi]{(corruption, template)} pairs, the proportion of times a given layer was in the minimal cross talk circuit. }
    \label{minimalcircuitproportionfigure}
\end{figure}

In Figure \ref{minimalcircuitproportionfigure}, we see that Layers 39 and 15 appear in every minimal circuit found. Layer 15 seems worthy of investigation in future work, as these two also stand out in \hyperref[appendix-eap]{EAP}. Inspecting the logs, Layer 39 is always the first layer added and has a large effect.

In 82\% of these minimal circuits, Layer 0 did not appear. This is strong evidence that for the majority of \hyperref[appendix-ioi]{(corruption, template)} pairs, computation Layer 0 does is in-place and not cross talk.

In Transformers, it is suspected that layer 0 is responsible for multi-token embeddings \citep{layer0detokenization}. These results suggest something else happens in Mamba. However, because all of our prompts use single token names, it is possible that these capabilities are simply not needed for this task (but still exist).

\subsection{Layer 39 Uses Conv to Shift Names One Position Forward}\label{layer39convshift}

When examining the hidden state, we can display the cosine similarity of a token's contribution to the current state with future (and previous) hidden states. This allows us to see how much the value was ``kept around'' (see \hyperref[appendix-ssm-hooks]{Appendix} for more information on the hooks used here).

\begin{figure}
    \includegraphics[width=\linewidth]{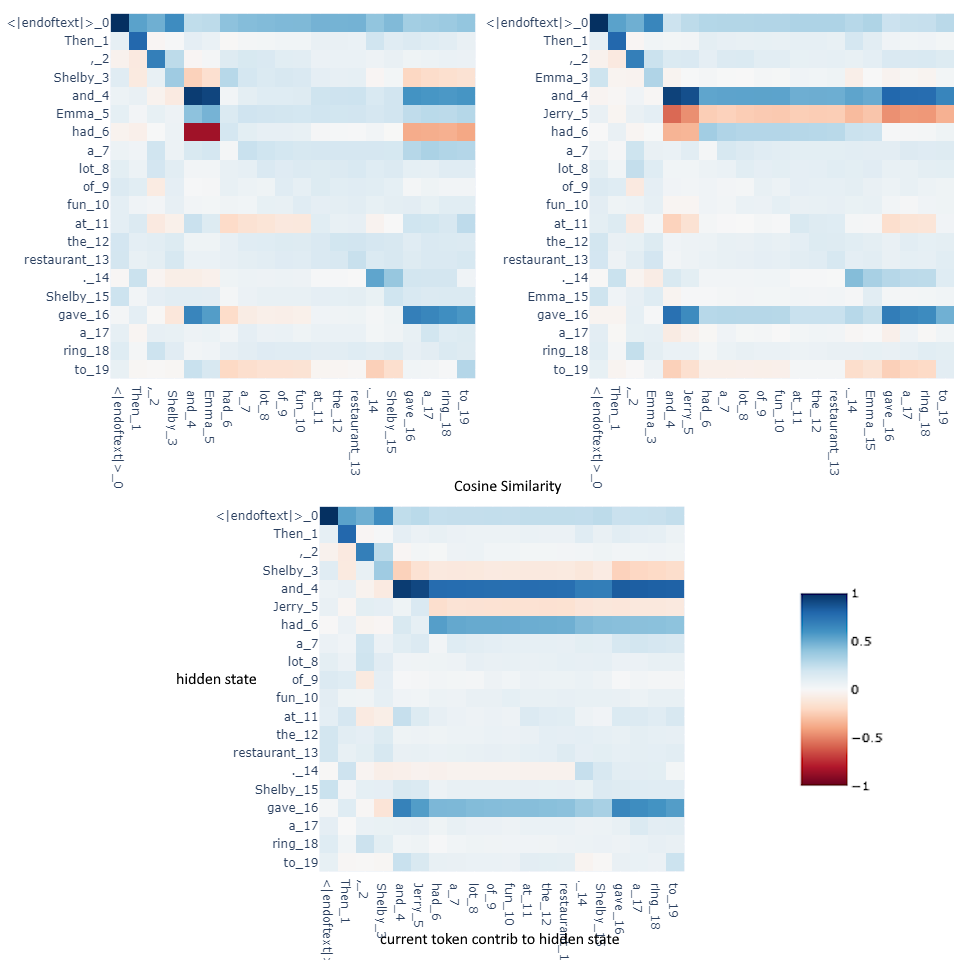}  
    \caption{Cosine Similarity between the current token's contribution to $h$ (which is $\overset{[B,E,N]}{\bar{B_i}} \overset{[B,E,1]}{x_i}$, each $i$ is on the x axis), and the hidden state ($\overset{[B,E,N]}{h_j}$ each $j$ is on the y axis)}
    \label{cosinesimfigure}
\end{figure}

As this is not causal, it should not be relied on too heavily. The structure seems to be name-dependent; we show three representative examples in Figure \ref{cosinesimfigure}.

What stands out is that the horizontal lines are one token after each name. This could either mean that 1) A previous layer shifted the tokens over, or 2) Layer 39 shifted the tokens over using the conv. 

To distinguish between these, we can do resample ablation on the individual conv ``slices'': The conv can be seen as four $E$-sized ``slice'' vectors for each (-3,-2,-1,0) relative token position, that are multiplied (element-wise) by the corresponding $E$-sized token representations.

\begin{itemize}
    \item If hypothesis 1 were true, we should see the 0 conv slice at token position + 1 have a large value.
    \item If hypothesis 2 were true, we should see the -1 conv slice at token position + 1 have a large value.
\end{itemize}

\begin{figure}
    \centering
    \includegraphics[width=1\linewidth]{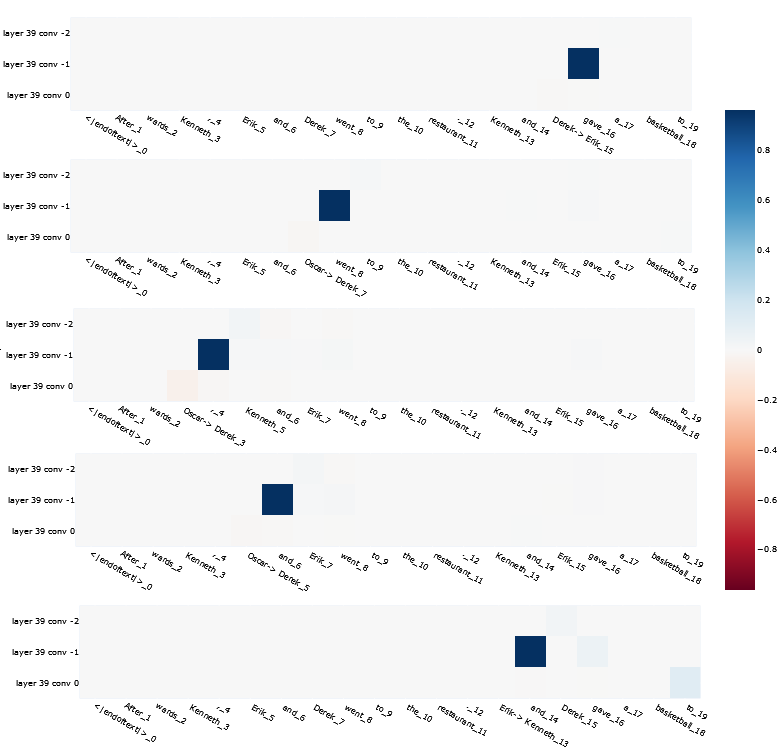}
    \caption{This is 1 - Normalized Logit Diff when patching on the given conv slice. 0 corresponds to acting like uncorrupted, 1 corresponds to acting like corrupted. The x-axis is conv slices (-2, -1, then 0) for layer 39. The y-axis is token position; observe the labels to see which corruption was used.}
    \label{convslicesablationfigure}
\end{figure}

Figure \ref{convslicesablationfigure} supports hypothesis 2, that Layer 39 uses conv to shift names one position forward.

We did some tests to investigate multi-token names and found the cosine similarity plots always have lines at the position after the first token of the name (possibly other layers handle multi-token names). It is also worth nothing that we see the horizontal lines for entities, not just names.

We do not yet know why this shifting behavior occurs, and leave that question for future work.

\subsection{Controlling Model Output by Modifying Representations on Layer 39}\label{overwritename}

We hypothesize that the representations in the SSM are linear because, on a single layer, the mechanism it has to add or remove information from tokens is linear in h.

We tried to visualize $\stackrel{[1]}{\Delta_{t,e}}$ adding or removing information to various parts but did not find it very insightful. Instead, to investigate whether the internal representation of Layer 39 SSM is linear, we do the following:

\begin{enumerate}
    \item Create a large IOI dataset. For each data point, store the activations of each name's \texttt{blocks.39.hook\_ssm\_input}. We use the activation at the token position one after the name, because of the shifting behavior we observed earlier.
    \item For each name, average the representations. Store a separate average for, say, ``John'' \hyperref[appendix-ioi]{in the first position}, ``John'' in the second position, etc. We use enough data points that each (name, position) pair gets 50-100 values to average over.
    \item \emph{Replace Method}: To write a different name, simply substitute the SSM input at that position with the averaged value from a different name.
    \item \emph{Subtract and Add Method} Instead of substituting, subtract the current name's average and add the substituted name's average.
\end{enumerate}

We find that the Replace Method works adequately, while the Subtract and Add Method works surprisingly well, changing the logits to the desired output more than 95\% of the time.



%

One thing to note: It was possible that the SSM was using the representations from the name's token position, as well as the name's token position + 1. Patching on conv slices was initial evidence this did not occur, and the efficacy of this replacement procedure provides further evidence that this is not the case.

Having a separate average for each position also lets us test if token position is an important part of the representation. If it is, we should expect that ``John'' at name position 2 should not be easily substituted for ``Mary'' at name position 0.

\begin{figure}
    \centering
    \includegraphics[width=1\linewidth]{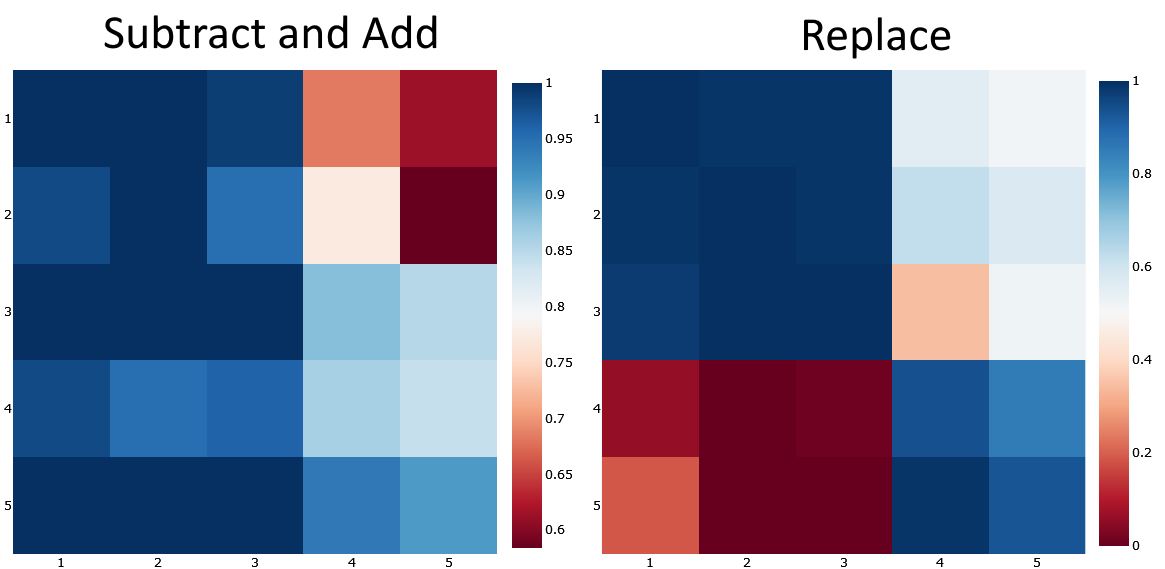}
    \caption{Proportion of data where logit of the corrupted name is higher than the logit of original name, using the two methods described in Section ~\ref{overwritename}. The x-axis is the position the average was computed from, the y-axis is the position being substituted. To substitute into the fourth and fifth positions, we substitute the correct answer (instead of a patched name).}
    \label{replacenamesfigure}
\end{figure}

Instead, in Figure \ref{replacenamesfigure} we find that the first three name positions are compatible, while the fourth and fifth positions are much less compatible.

\subsubsection{Compatibility of first three name positions}

The compatibility of the first three name's representations could either suggest:

\begin{itemize}
    \item The IOI circuit does not store positional information in the first three names, or
    \item There are circuits to handle incorrectly encoded positional information, which got activated and handled our patching well despite having incorrect positional data
\end{itemize}

Distinguishing between these is left for future work.

\subsubsection{Incompatibility of fourth and fifth name positions}

Consider one of our data points: ``Friends Isaac, Lucas and Lauren went to the office. Isaac and Lucas gave a necklace to'' (answer is `` Lauren'')

We see that when names occur in the fourth or fifth position, it is the second time they occur in the prompt.

One hypothesis is that the conv sees a period and encodes that in the name representation. However, while the model we study (mamba-370m) has four conv slices, we find that the conv slice attending to the -3 position is always zero, likely due to a bug in the Mamba training code. Thus, it can only attend to the previous 2 positions in practice (the third conv slice is for attending to the current position). This means that the name in the fifth position's representation is not distinguishable from the name in the third position by the conv.

Thus, some token cross-talk must be happening in a layer before 39. As mentioned above, for 82\% of (corruption, template) pairs, cross-talk in layer 0 is not needed. So while these experiments provide strong evidence that layer 39 is a bottleneck, more circuit analysis is needed.

\subsection{Layer 39 moves information into only the last token position}\label{layer39moveintolasttokenpos}

We can do resample ablation on the ssm hidden state via \texttt{blocks.\{layer\}.hook\_h.\{token\_pos\}}.

Figure \ref{hresampleablationfigure} shows that it only uses the hidden states one after the ablated token, in line with ~\ref{layer39convshift}. We also see that hidden state values are used all the way to the last token position.

This tells us that the answer-relevant information is moved into the last token position. However, it is possible that information is also sent to other, earlier positions as well.

To test for this, we can do resample ablation on $\texttt{blocks.\{layer\}.hook\_proj\_out}$, which is the value added to the residual stream at the end of each layer.

In Figure \ref{projoutresampleablation} we see that only the last index is used.

In addition, positional EAP (Section \ref{from39connections}) suggests that other (non-last token) connections are important, as they are preserved in the set of edges that get 85\%. However their attribution scores are very low. Manually removing all the non-last token connections going out from layer 39 only reduces accuracy from 85.2\% to 83.8\%, and reduces normalized logit diff from to 0.877 to 0.873. This suggests that either there is backup behaviour activated when those positions are patched, or that these connections are mostly spurious and not essential parts of the circuit.

These three lines of evidence together strongly suggest that the task-relevant information provided by layer 39 is stored only in the last token position.

\section{Positional Edge Attribution Patching (Positional EAP)}

Here we describe a simple modification to EAP that allows us to have token-level edge attributions.

Typically, in EAP, after we compute $\overset{[B,L,D]}{\text{attr}_{i\mapsto j}}$ we \texttt{sum} over the $L$ and $D$ dimensions, then take the \texttt{mean} over the $B$ dimension to get an attribution for each edge.

Instead, we will just \texttt{sum} over the $D$ dimension and \texttt{mean} over the $B$ dimension, giving us an attribution for every (edge, position). See Appendix ~\ref{appendix-eap}.

The results of EAP further emphasize the importance of Layer 39. However, there is also significant activity elsewhere that merit further analysis.

\section{Future Work}

There are still many open questions we have about the IOI circuit in mamba-370m. Future work can focus on:

\begin{itemize}
    \item Analysis of what cross talk is done before layer 39
    \item Analysis of what the later layers are doing to decode the answer encoded in the final token position
    \item Training Sparse Autoencoders (SAEs) and using EAP to make a feature circuit capable of doing the task, to get a more fine grained analysis (similar to work in \citet{marks2024sparse})
    \item Conducting similar analysis on other tasks (such as docstring \citep{docstring} or greater than \citep{greaterthan})
\end{itemize}

\section{Reproducibility}

All code for experiments can be found at \href{https://github.com/Phylliida/investigating-mamba-ioi}{https://github.com/Phylliida/investigating-mamba-ioi}. All experiments were conducted on a RTX A6000.

\ifdefined\isaccepted
\section{Credits}

The authors would like to thank the ML Alignment \& Theory Scholars (MATS) program for providing a workspace to conduct this research, FAR AI Labs for compute, and LTFF for funding. We would also like to thank Niels uit de Bos, Iván Arcuschin Moreno, 
Rohan Gupta, Thomas Kwa, Scott Neville, Gonçalo Paulo and Joseph Bloom for the helpful conversations.
\else
\fi

\begingroup
\RaggedRight
\bibliography{sample}
\bibliographystyle{icml2024}
\endgroup

\begin{appendices}

\begin{figure*}[hbtp]
    \centering
    \includegraphics[width=0.7\linewidth]{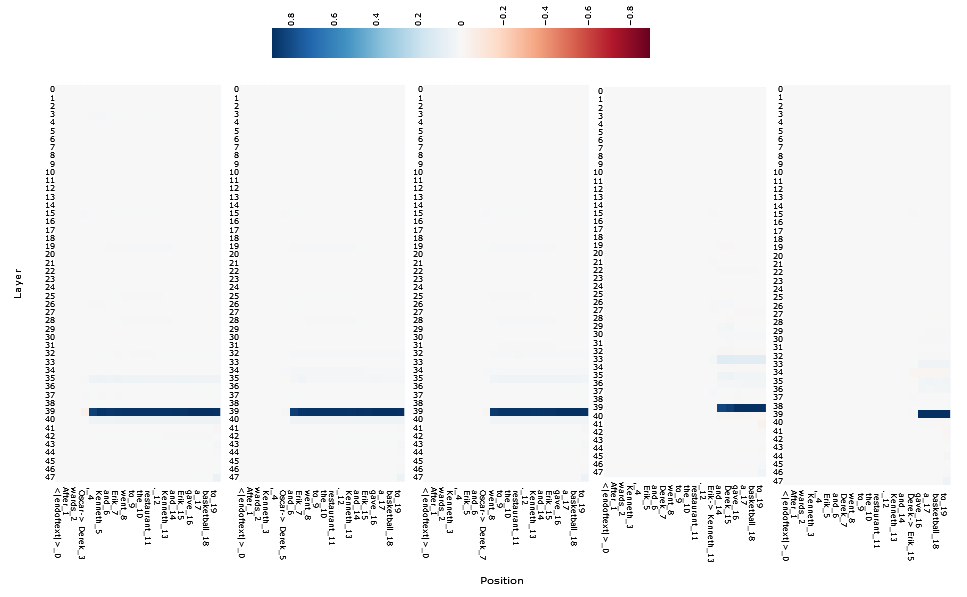}
    \caption{ Displayed is 1 - (Normalized logit diff) for each (layer, position) patch, averaged over 80 data points. 0 corresponds to acting like the uncorrupted forward pass, and 1 corresponds to acting like the corrupted forward pass. The y-axis is Layer, and the x-axis is token position. The corruptions can be observed by inspecting the token position labels. Each of the five plots correspond to different IOI patches.}
    \label{hresampleablationfigure}
\end{figure*}

\begin{figure*}[hbtp]
    \centering
    \includegraphics[width=0.7\linewidth]{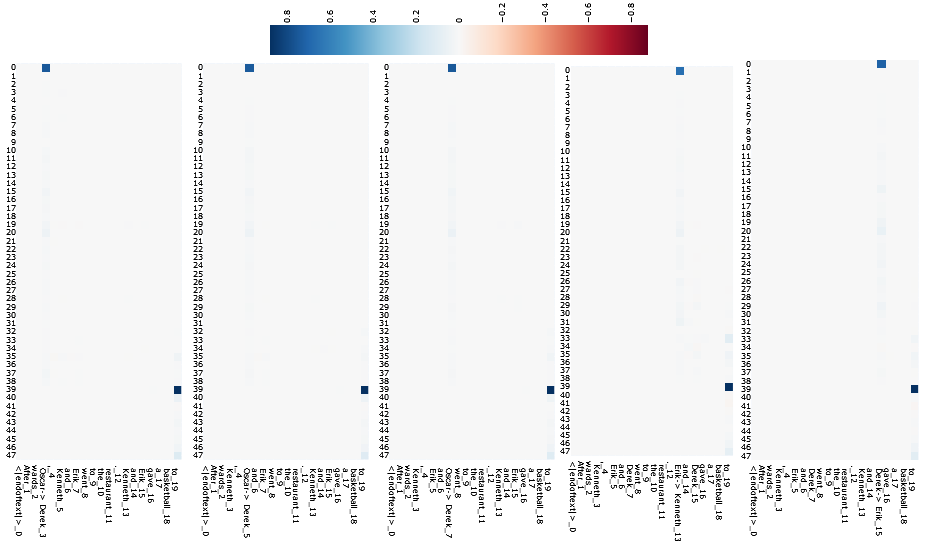}
    \caption{ Same as Figure \ref{hresampleablationfigure}, but for $\texttt{blocks.\{layer\}.hook\_proj\_out}$}
    \label{projoutresampleablation}
\end{figure*}

\section{IOI Task Details}\label{appendix-ioi}

We use the 4 prompt templates from \citep{conmy2023automated}:

\begin{verbatim}
Then, [NAME], [NAME] and [NAME] went to
the [PLACE]. [NAME] and [NAME] gave a
[OBJECT] to

Afterwards [NAME], [NAME] and [NAME]
went to the [PLACE]. [NAME] and [NAME]
gave a [OBJECT] to

When [NAME], [NAME] and [NAME] arrived
at the [PLACE], [NAME] and [NAME] gave
a [OBJECT] to

Friends [NAME], [NAME] and [NAME] went
to the [PLACE]. [NAME] and [NAME] gave
a [OBJECT] to
\end{verbatim}

In resample ablation, there are many ways to corrupt a prompt. We create a dataset choosing randomly from all possible corruptions and locations of names that:

\begin{enumerate}
\item Replace all instances of a single name with another name, and
\item Change the output
\end{enumerate}

While we could patch two names at the same time, 1 simplifies the number of things being changed at the same time. 2 is necessary to determine if the patch had any effect.

This results in the following 5 corruptions:

\begin{verbatim}
CAB AB C
DAB AB D

ACB AB C
ADB AB D

ABC AB C
ABD AB D

ABC AB C
ABC AC B

ABC AC B
ABC BC A
\end{verbatim}

In each of these, the top line represents the uncorrupted prompt, and the bottom line represents the corrupted prompt. Letters correspond to names: the first three are the first three names, the second two are the fourth and fifth names, and the last is the output. If two letters are the same, that means that those places share the same name. Otherwise, the names are different.

\section{Normalized Logit Diff}\label{appendix-metric}

Normalized Logit Diff is defined as:

\begin{verbatim}
min_diff =
    A_logit_corrupted - B_logit_corrupted
max_diff =
    A_logit_unpatched - B_logit_unpatched
possible_range = abs(max_diff - min_diff)
# prevent divide by zero
possible_range[possible_range == 0] = 1
logit_diff =
    A_logits_patched - B_logits_patched
normalized_logit_diff =
    (logit_diff-min_diff)/possible_range
\end{verbatim}

where

\begin{itemize}
    \item unpatched is a forward pass without our intervention (the baseline forward pass)
    \item corrupted is a forward pass without our intervention, where the prompt is modified to make $B$ the correct answer
    \item patched is a forward pass that patches edges (as described above)
\end{itemize}

This results in a 1 when the model acts like the unpatched forward pass, and 0 when the model acts like the corrupted forward pass. Note that it is possible to obtain scores outside the [0,1] range.

The abs is a novel addition by us. For data points where the model is incorrect, maximizing normalized logit diff would result in the model becoming more incorrect. This abs modification fixes that issue.

\section{SSM Hooks used}\label{appendix-ssm-hooks}

For the cosine similarity plots, we used \texttt{blocks.39.hook\_B\_bar}, \texttt{blocks.39.hook\_ssm\_input}, and  \texttt{blocks.39.hook\_h.\{pos\}}
    
\begin{figure*}[t]
\includegraphics[width=\linewidth]{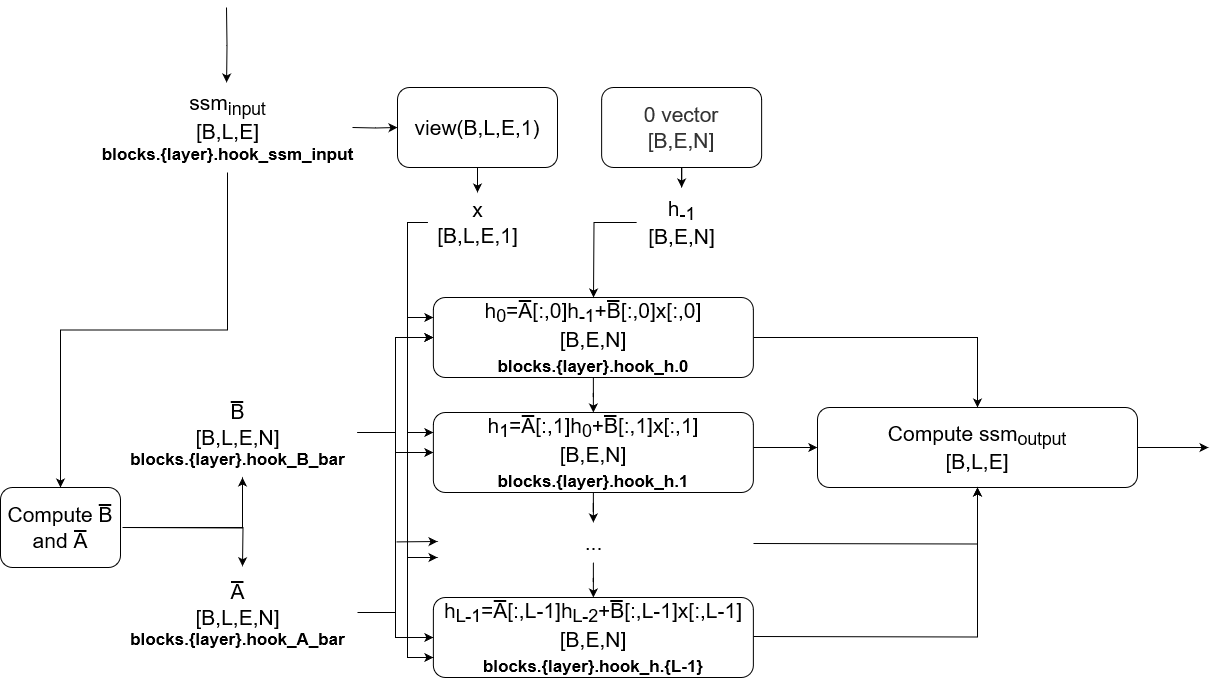}
\label{figure-ssm-internals}
\caption{Internals of the SSM block, we restrict this diagram to only the parts we are interested in}
\end{figure*}

See figure ~\ref{figure-ssm-internals} for a detailed overview of the SSM nternals.

\section{Automated Circuit Discovery Results}\label{appendix-eap}

We use the following edges:

\begin{itemize}
    \item embed $\mapsto$ layer input (hook\_embed $\mapsto$ blocks.i.hook\_layer\_input)
    \item layer output $\mapsto$ later layer input (blocks.i.hook\_proj\_out $\mapsto$ blocks.j.hook\_layer\_input)
    \item layer output $\mapsto$ output (blocks.i.hook\_proj\_out $\mapsto$ blocks.47.hook\_resid\_post)
    \item embed $\mapsto$ output hook\_embed $\mapsto$ blocks.47.hook\_resid\_post)
\end{itemize}

Where output is the residual stream after the final layer has added its layer output.

We do a few separate experiments.

\subsection{EAP}
    
At the most high level, we can run (integrated gradient) EAP without positions. Using binary search to find the minimum number of edges to give us at least 85\% accuracy results in the following adjacency matrix:

\begin{figure}[H]
    \centering    \includegraphics[width=1\linewidth]{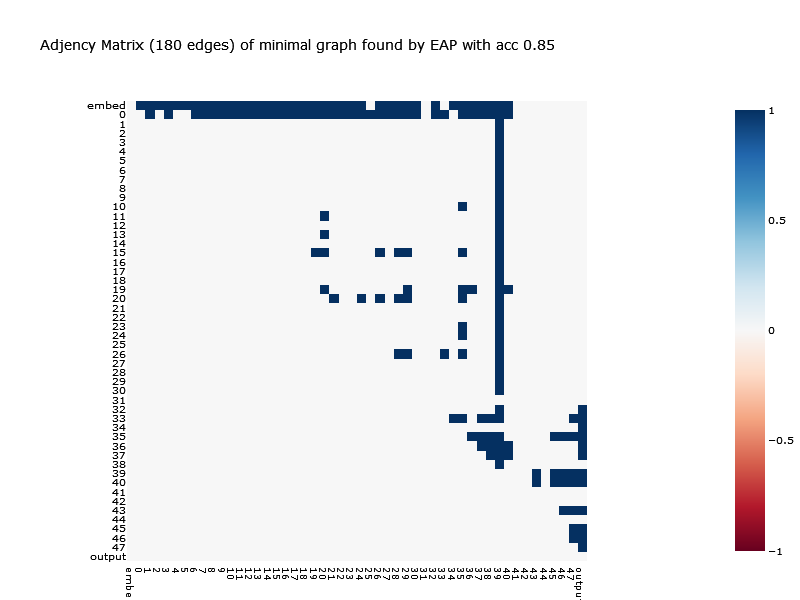}
    \caption{Integrated gradients EAP, minimum set of edges. A blue dot means the edge is present. The y-axis is the input node, the x-axis is the output node}
    \label{fig:nopositionseap}
\end{figure}

We also present the edge attributions here, and the corresponding actual effects on normalized logit diff (determined during the ACDC run below)

\begin{figure*}[hbtp]
  \centering
  \begin{minipage}[b]{0.4\textwidth}
    \includegraphics[width=\textwidth]{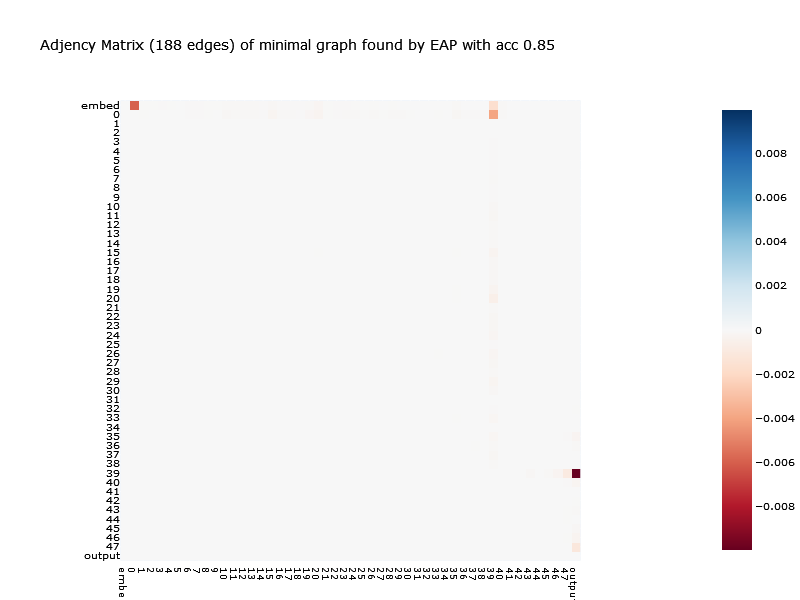}
    \caption{Edge Attributions}
  \end{minipage}
  \hfill
  \begin{minipage}[b]{0.4\textwidth}
    \includegraphics[width=\textwidth]{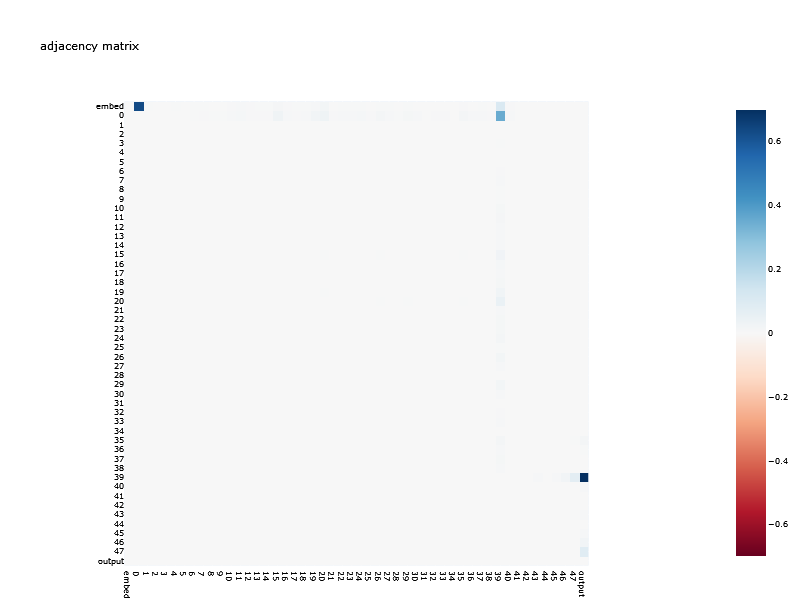}
    \caption{Effects on normalized logit diff}
  \end{minipage}
\end{figure*}

This is not very insightful, so we clamp values to let us see more (the thresholds chosen by hand)

\begin{figure*}[hbtp]
  \centering
  \begin{minipage}[b]{0.4\textwidth}
    \includegraphics[width=\textwidth]{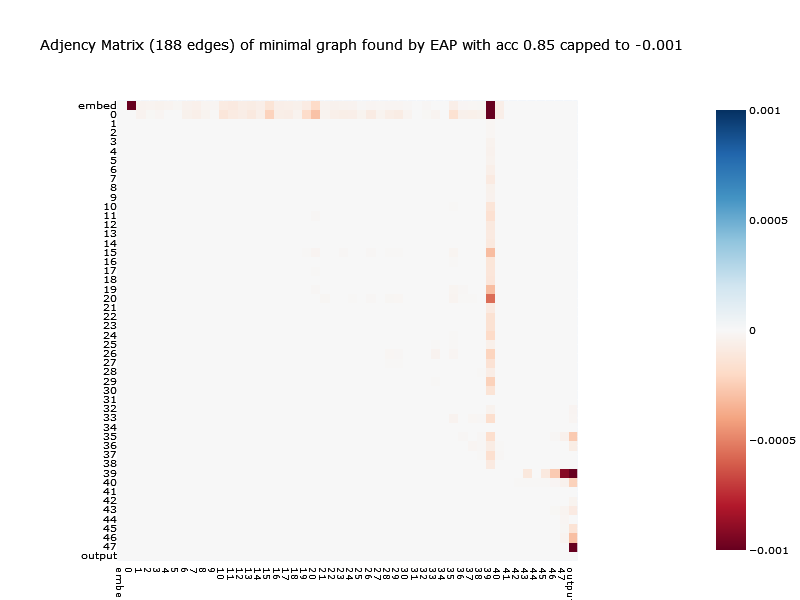}
  \end{minipage}
  \hfill
  \begin{minipage}[b]{0.4\textwidth}
    \includegraphics[width=\textwidth]{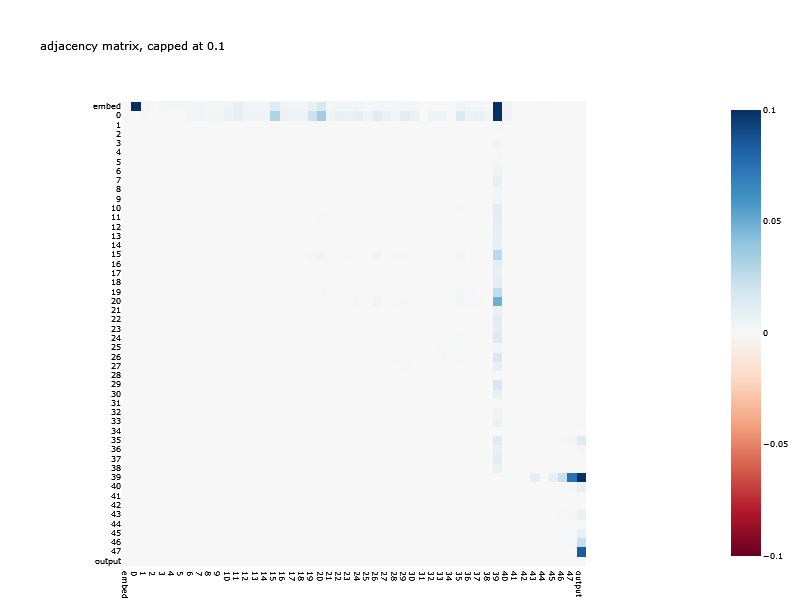}
  \end{minipage}
\end{figure*}

\begin{figure*}[hbtp]
  \centering
  \begin{minipage}[b]{0.4\textwidth}
    \includegraphics[width=\textwidth]{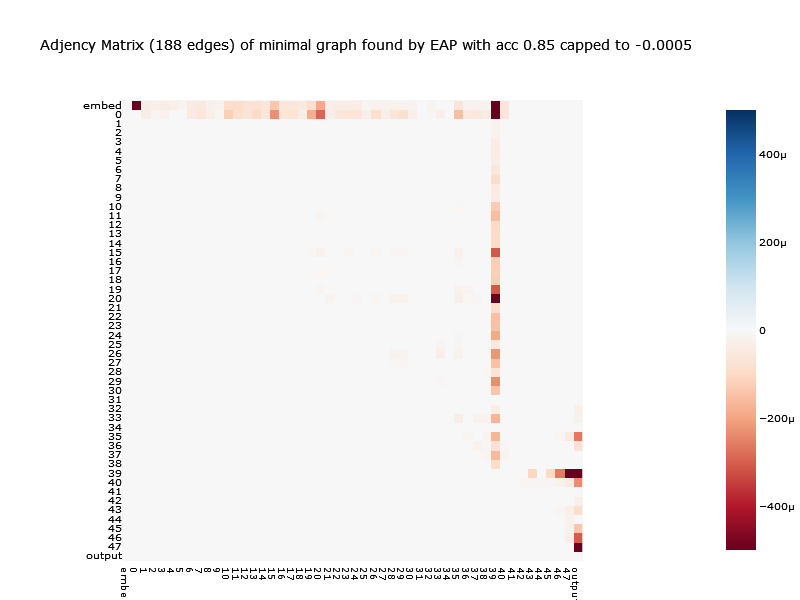}
  \end{minipage}
  \hfill
  \begin{minipage}[b]{0.4\textwidth}
    \includegraphics[width=\textwidth]{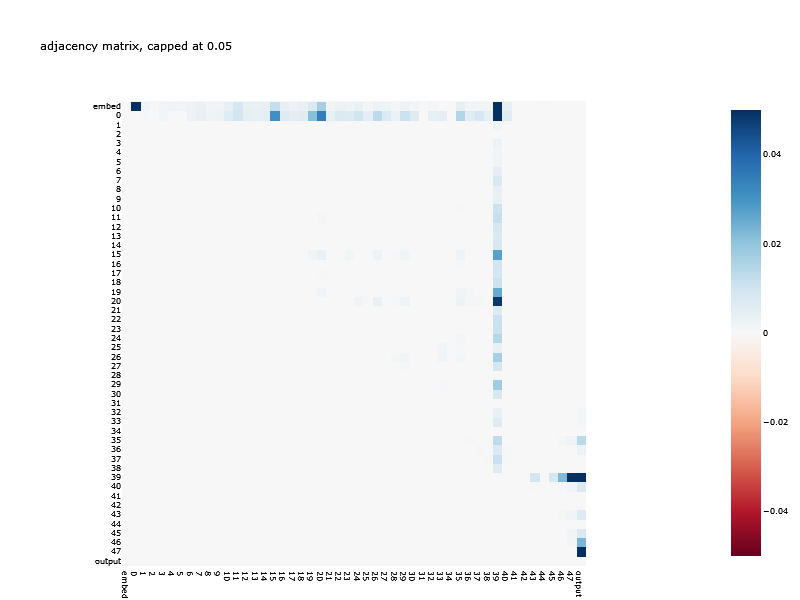}
  \end{minipage}
\end{figure*}

\begin{figure*}[hbtp]
  \centering
  \begin{minipage}[b]{0.4\textwidth}
    \includegraphics[width=\textwidth]{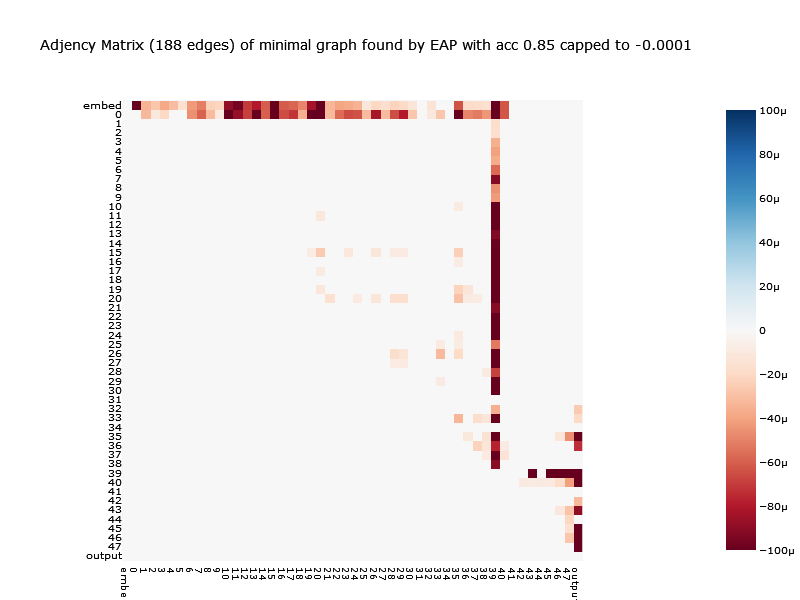}
  \end{minipage}
  \hfill
  \begin{minipage}[b]{0.4\textwidth}
    \includegraphics[width=\textwidth]{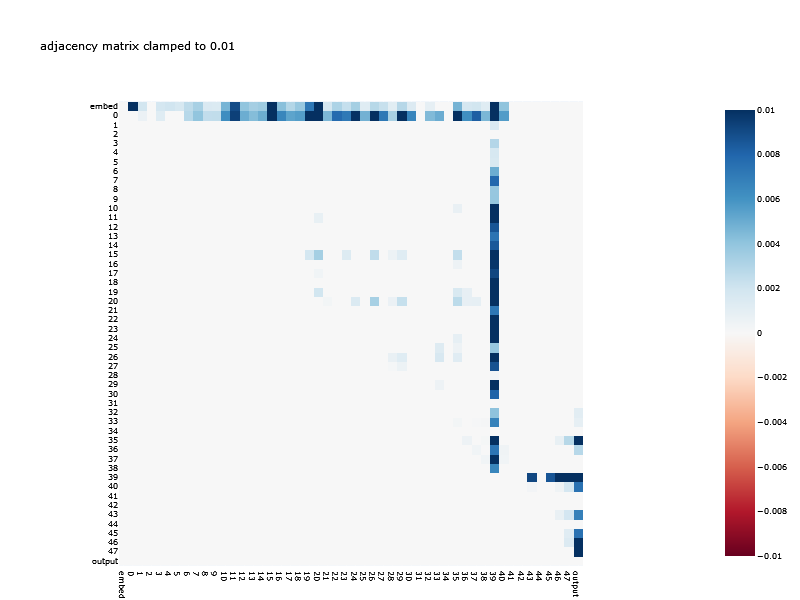}
  \end{minipage}
\end{figure*}

\subsection{ACDC for layer information}\label{ACDC}

We run EAP without positions using the edges listed above, and use binary search to find the least edges needed to get 85\% accuracy. We take the resulting graph and run ACDC on it with a thresh of 0.0001, with (non-positional) edges for individual conv slices, the ssm, the skip connection, and all the edges above.

This allows us to get a hint at what parts each layer is using.

\begin{figure*}[hbtp]
    \centering
    \includegraphics[width=1\linewidth]{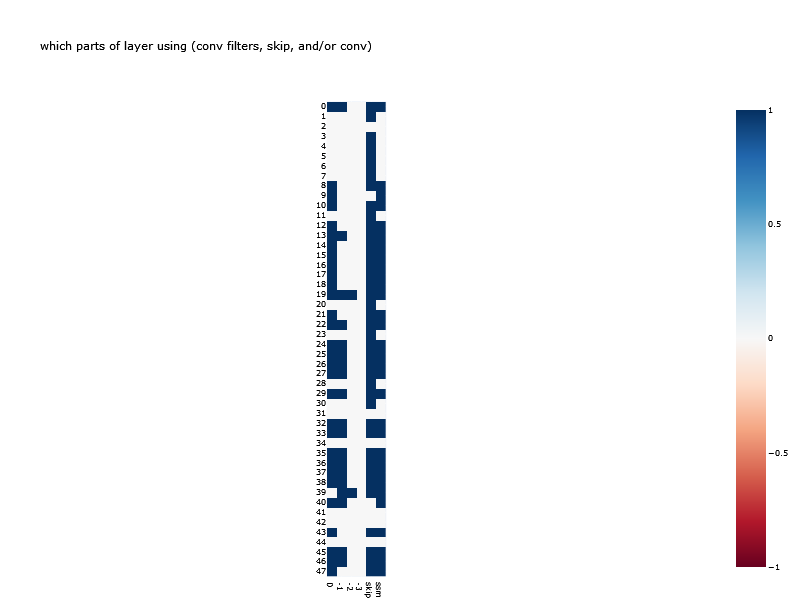}
    \caption{ACDC results from inside layers. Each node is 1 if the edge is present, 0 if it is not. 0,-1, and -2 are the corresponding conv slices}
    \label{fig:whichpartsacdc}
\end{figure*}

\begin{figure*}[hbtp]
    \centering
    \includegraphics[width=1\linewidth]{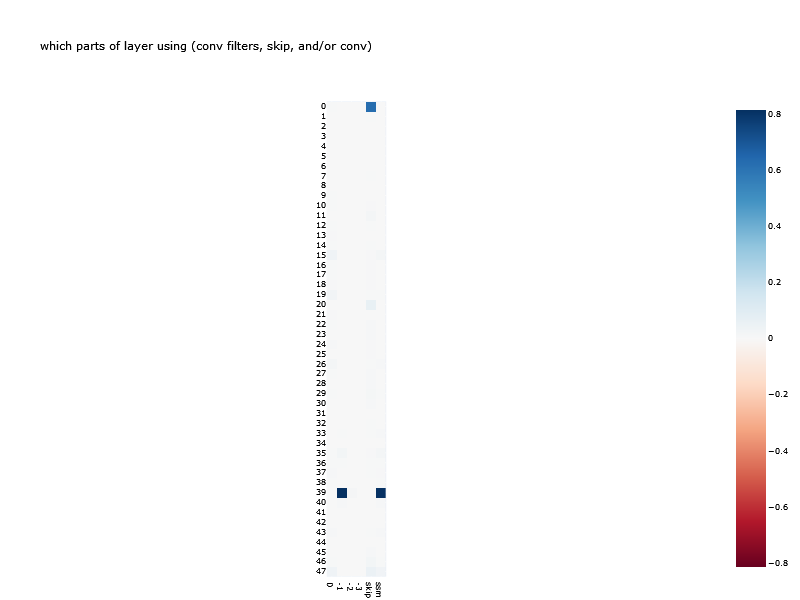}
    \caption{Same as the above figure, however, each cell shows the decrease in normalized logit diff if that edge is patched}
    \label{fig:whichpartssmooth}
\end{figure*}

\begin{figure*}[hbtp]
    \centering
    \includegraphics[width=1\linewidth]{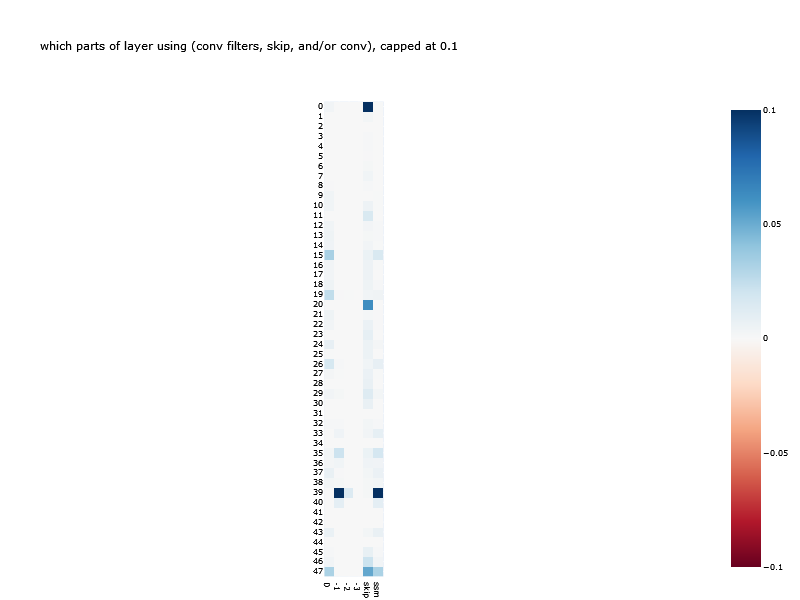}
    \caption{Same as above, however, values are clamped to 0.1}
    \label{fig:whcihpartssmoothsmaller-label}
\end{figure*}

\begin{figure*}[hbtp]
    \centering
    \includegraphics[width=1\linewidth]{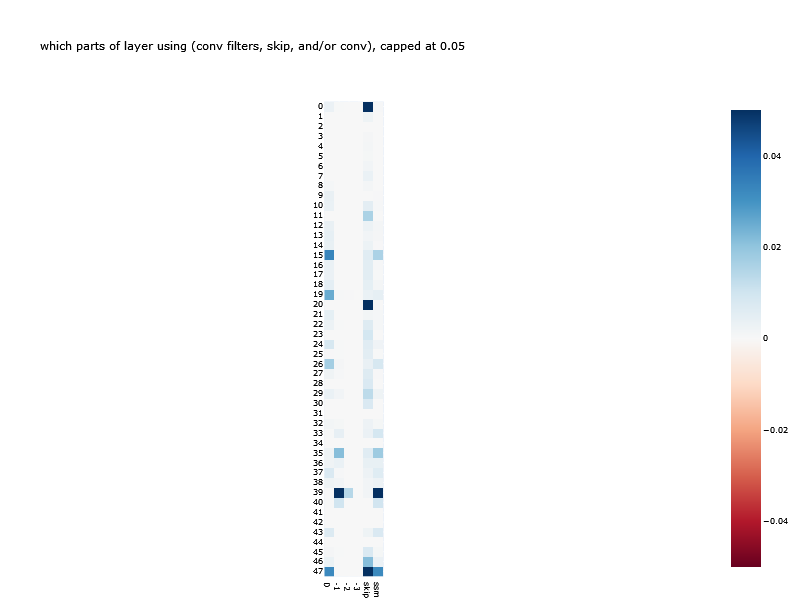}
    \caption{Same above, however, values are clamped to 0.05}
    \label{fig:whichpartssmoothsmallerrer-label}
\end{figure*}

This resulting circuit has an average normalized logit diff of 0.84 and achieves 88\% accuracy on a held-out test set, so there is little loss in performance from doing this further prune (because the thresh is so low, most edges pruned are those that decrease ability to do the task, which is why accuracy has gone up). 

Of note, this reproduces the ``Convs of layer 39 shift names to the next position'' result from above.

Ideally we could do this with EAP and no longer need ACDC, but but leave that for future work.

\subsection{EAP With Positions}

In the following:

\begin{itemize}
    \item n1 means the first name in the prompt, n2 means the second name in the prompt, etc.
    \item pos0 means the first token, pos1 means the second token, etc. (these are used for non-name tokens)
    \item out means the final token, where the answer is generated
\end{itemize}

For reference, here is a prompt:

\begin{verbatim}
pos0 <|endoftext|>
pos1 Then
pos2 ,
n1 Sally
pos4 ,
n2 Martha
pos6 and
n3 Edwin
pos8 went
pos9 to
pos10 the
pos11 restaurant
pos12 .
n4 Edwin
pos14 and
n5 Sally
pos16 gave
pos17 a
pos18 drink
out to
\end{verbatim}

\subsubsection{Connections from embed}

Every layer receives n1-n5, except:
\begin{verbatim}
Missing n1: layers 1, 5, 8, 25, 28, 29,
     30, 32, 34, 36
Missing n2: 30, 31, 36
Missing n3: layers 25, 27, 29, 30, 31,
     34, 36, 38
Missing n4: layers 3, 9, 25, 32, 34, 40
Missing n5: layer 9, 25, 27, 31, 34, 37,
     38, 40
Missing n1-n5: 33, 41-47
TODO: Output?
\end{verbatim}

\subsubsection{Connections of Layer 0}

Layer 0 takes as input n1-n5, and sends n1-n5 to every layer, except:
\begin{verbatim}
Missing n1: 1, 2, 3, 28, 31, 32, 33, 34
Missing n2: 6, 7, 31, 33
Missing n3: 2, 31, 33
Missing n4: 1, 2, 3, 9, 18, 21, 32, 34, 40
Missing n5: 3, 9, 18, 31, 32, 34, 40
Missing n1-n5: 4, 5, 41-47
\end{verbatim}

\subsubsection{Layers that are missing names}

If we consinder a layer as ``having'' a name if it received it from embed or layer 0, the following layers have n1-n5:

And these are ones that are missing names:

\begin{verbatim}
Missing n1: 1, 5, 28, 32, 33, 34
Missing n2: 31, 33
Missing n3: 31, 33
Missing n4: 32, 34, 3, 40, 9
Missing n5: 9, 31, 34, 40
Missing n1-n5: 41-47
\end{verbatim}

Of those, \texttt{1, 3, 5, 9} are only connected to 0/embed and 39. In particular, we have:

\begin{verbatim}
1: missing n1
   n1,n2,n3,n5 -> 39
3: missing n4
   n1-n5 -> 39
5: missing n1
   n1-n5 -> 39
9: missing n4,n5
   n1-n5 -> 39
\end{verbatim}

Otherwise, we have

\begin{verbatim}
Missing n1: 28, 32, 33, 34
Missing n2: 31, 33
Missing n3: 31, 33
Missing n4: 32, 34, 40
Missing n5: 31, 34, 40
Missing n1-n5: 41-47
\end{verbatim}

\texttt{28, 31, 32, 33, 34, 40} all seem to be involved in a complex circuit, and have inputs from other layers.

Just examining the missing terms:

\begin{figure}[H]
    \centering    \includegraphics[width=0.75\linewidth]{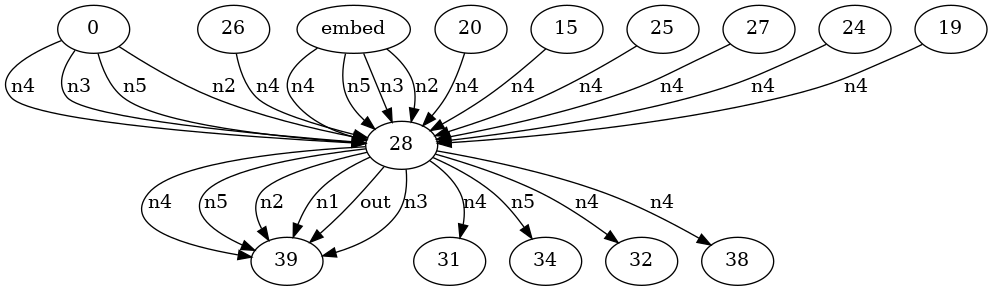}
\end{figure}

\begin{verbatim}
28 is missing n1

28 does not receive n1 from anyone
    outputs n1 to 39
\end{verbatim}

\begin{figure}[H]
    \centering    \includegraphics[width=0.75\linewidth]{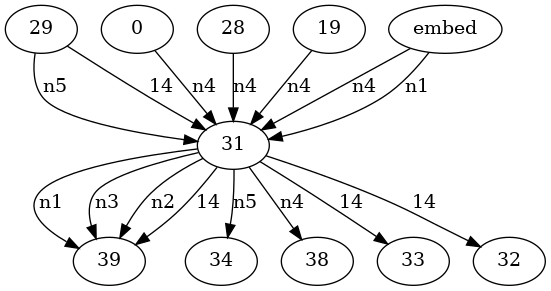}
\end{figure}

\begin{verbatim}
31 is missing n2, n3, n5

31 does not receive n2 or n3 from anyone
    output n2-n3 to 39
31 receives n5 from 29
    outputs n5 to 34
\end{verbatim}

\begin{figure}[H]
    \centering    \includegraphics[width=0.75\linewidth]{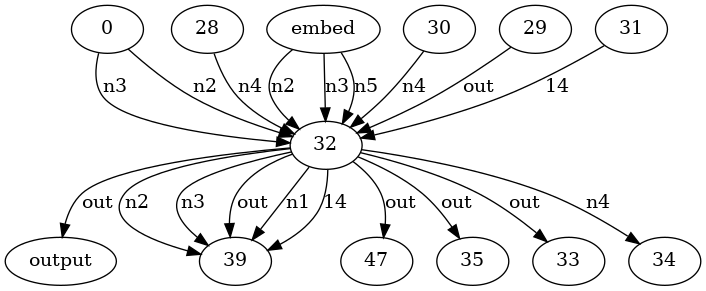}
\end{figure}

\begin{verbatim}
32 is missing n1, n4

32 does not receive n1
   outputs n1 to 39
32 receives n4 from 28, 30
   outputs n4 to 34
\end{verbatim}

\begin{figure}[H]
    \centering    \includegraphics[width=0.75\linewidth]{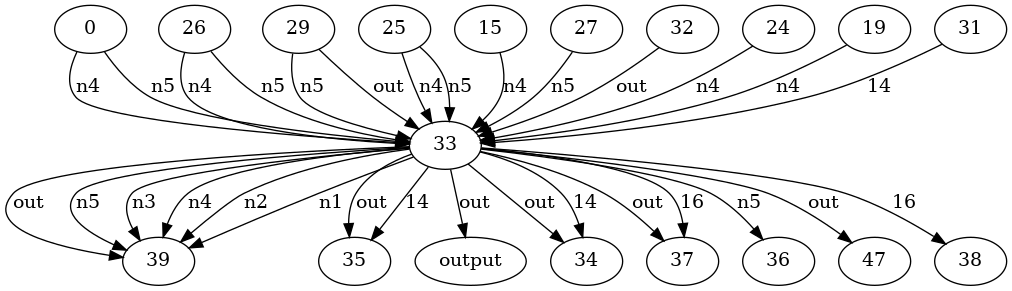}
\end{figure}

\begin{verbatim}
33 is missing n1-n3

33 does not receive n1-n3
   outputs n2-n3 to 39
   outputs n1 to 35
\end{verbatim}

\begin{figure}[H]
    \centering    \includegraphics[width=0.75\linewidth]{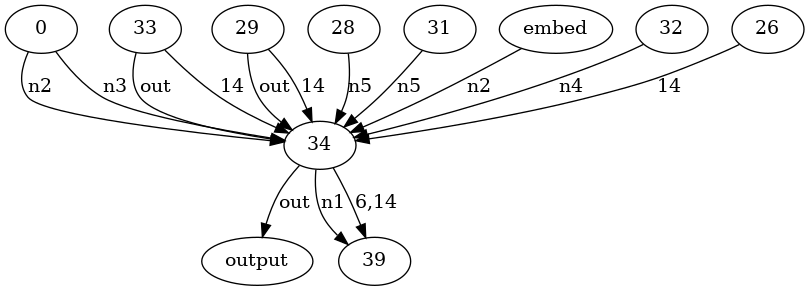}
\end{figure}

\begin{verbatim}
34 is missing n1, n4, n5

34 does not receive n1 from anyone
    outputs n1 to 39

34 receives n4 from 32
    does not output it

34 receives n5 from 28, 31
    does not output it
\end{verbatim}

\begin{figure}[H]
    \centering    \includegraphics[width=0.75\linewidth]{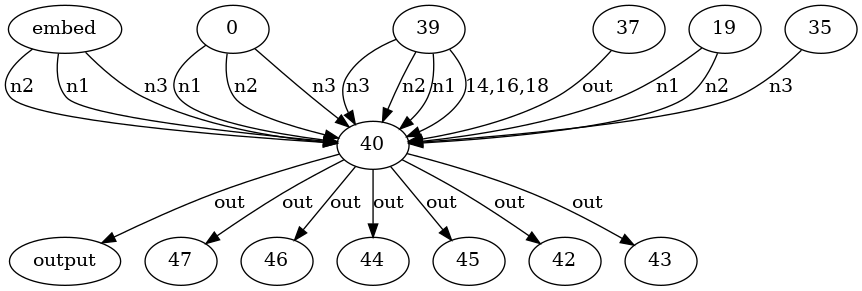}
\end{figure}

\begin{verbatim}
40 is missing n4, n5

40 does not receive n4-n5
    does not output n4-n5
\end{verbatim}

\subsubsection{Connections to 39}

As expected, layer 39 stands out as noteworthy. Every layer before 39 has a connection to 39 for every name, with these exceptions:

\begin{verbatim}
Missing n1: layer 2
Missing n2: layer 34
Missing n3: layer 34
Missing n4: layers 1, 31, 32, 34
Missing n5: layers 2, 31, 32, 34
\end{verbatim}

In addition, there are these extra connections to layer 39
\begin{verbatim}
pos6: layer 34
pos14: layers 31, 32, 34
out: layers  28, 29, 30, 33, 35, 37, 38
\end{verbatim}

Where

\begin{itemize}
    \item pos6 is the `` and'' between n2 and n3
    \item pos14 is the “ and” between n4 and n5
\end{itemize}

\subsubsection{Connections from 39}\label{from39connections}

\begin{verbatim}
n1,n2,n3: layer 40
pos12: layer 43
pos14: layer 40
pos16: layers 40,41,43,44,45,46,47
pos18: layer 40
out: layers 43,45,46,47, and output
\end{verbatim}

\begin{itemize}
    \item pos12 is the ``.''
    \item pos14 is the `` and'' between n4 and n5
    \item pos16 is the `` gave'' after n5
    \item pos18 is the object (for example, `` drink'')
\end{itemize}

\subsubsection{Graph after hiding 39, embed, and 0}

Keeping the above in mind, once we hide those three nodes the graph is quite readable. If we hide 35, it is even more readable. We also plot layer 35, for reference.

\begin{figure*}[t]
\includegraphics[width=1.1\linewidth]{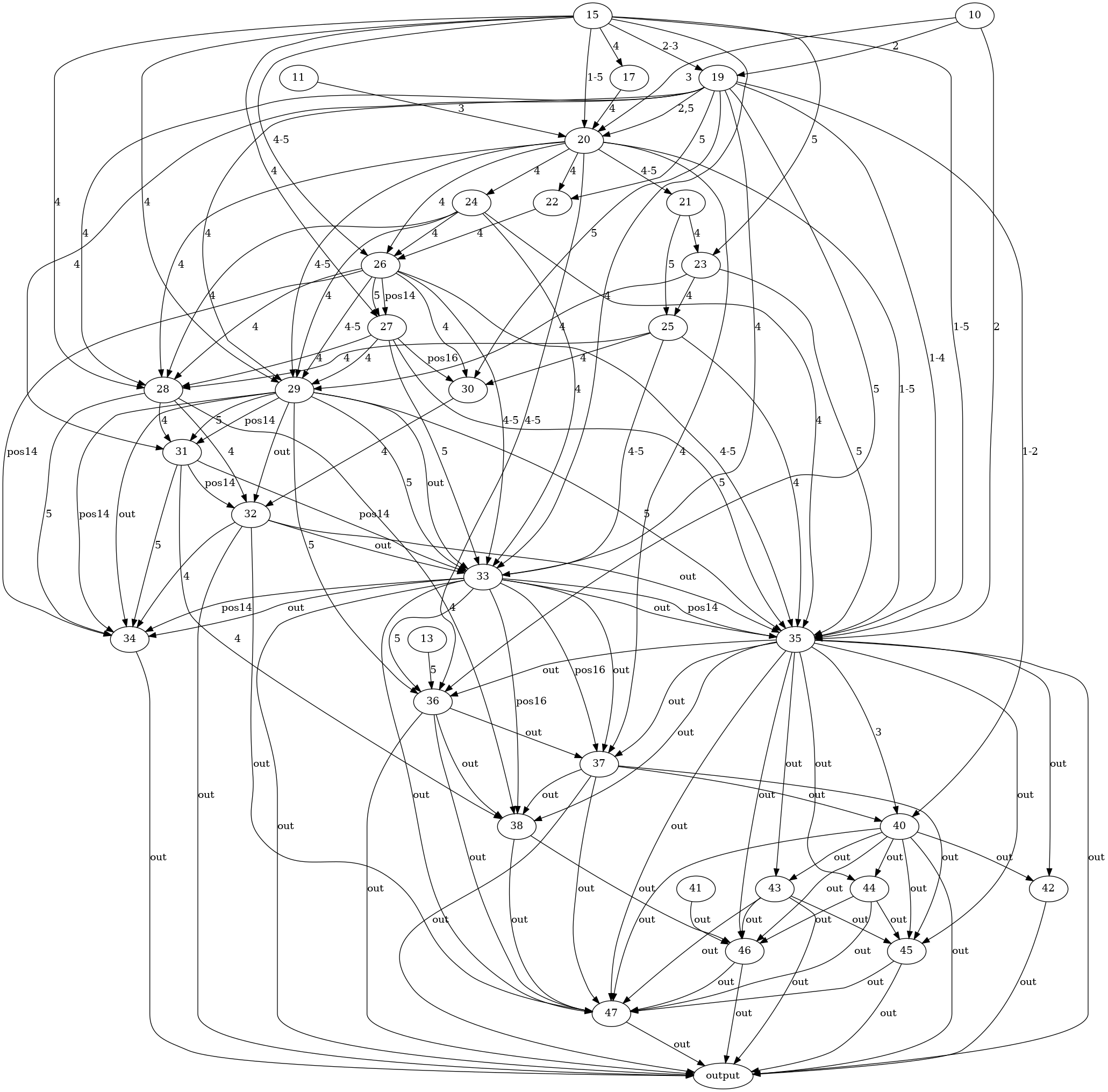}
\label{eap pos resultsffff}
\caption{Positional EAP After hiding Embed, Layer 0, and Layer 39. Numbers correspond to names, pos14 means token in position 14, out means the final token}
\end{figure*}

\begin{figure*}[t]
    \centering
    \includegraphics[width=1.1\linewidth]{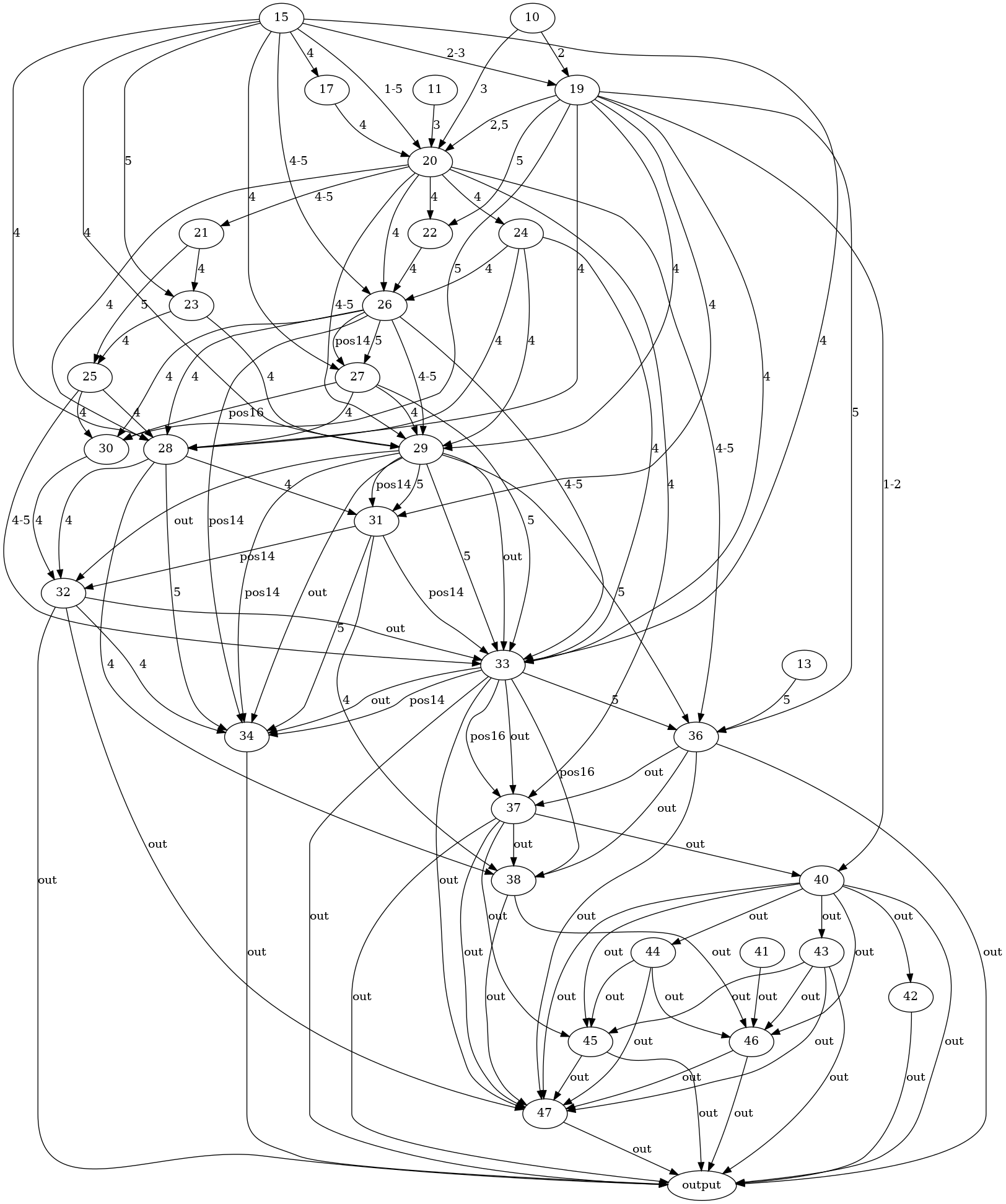}
    \caption{Same as above, but also hide 35}
\end{figure*}

\begin{figure*}[t]
    \centering
    \includegraphics[width=1\linewidth]{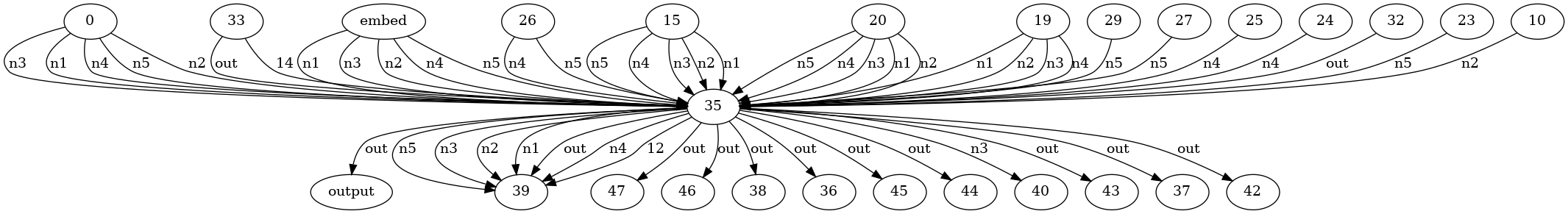}
    \caption{Layer 35 Positional EAP results}
    \label{fig:35layerviewfig}
\end{figure*}

\end{appendices}

\end{document}